\documentclass[runningheads]{llncs}
\usepackage[T1]{fontenc}
 \usepackage{amsmath}
\usepackage{graphicx}
\usepackage{orcidlink}
\usepackage{xcolor}
\usepackage{cleveref}

\begin{document}
\title{Examining the legibility of humanoid robot arm movements in a pointing task}
\titlerunning{Legibility of robot arm movements}
\author{Andrej L\'u\v{c}ny\inst{1}$^{\orcidlink{0000-0001-6042-7434}}$ \and
Matilde Antonj\inst{2,3}$^{\orcidlink{0000-0003-2500-7754}}$ \and 
Carlo Mazzola\inst{3, 4}$^{\orcidlink{0000-0002-9282-9873}}$ \and 
Hana Horn\'a\v{c}kov\'a\inst{1}
Ana Fari\'c\inst{5} \and 
Krist\'ina Malinovsk\'a\inst{1}$^{\orcidlink{0000-0001-7638-028X}}$ \and
Michal Vavre\v{c}ka\inst{1}$^{\orcidlink{0000-0002-0152-2682}}$ \and
Igor Farka\v{s}\inst{1}$^{\orcidlink{0000-0003-3503-2080}}$
}
\authorrunning{A. L\'u\v{c}ny et al.}
%
\institute{Faculty of Mathematics, Physics and Informatics, \\ Comenius University Bratislava, Slovakia\\
\and
DIBRIS, University of Genoa, Genoa, Italy
\and
CONTACT, Italian Institute of Technology, Genoa, Italy\\
\and
Digital Health Dept., Nvision Systems \& Technologies, S.L., Barcelona, Spain\\
\and
Faculty of Education, University of Ljubljana, Slovenia
}
\maketitle   

\begin{abstract}
Human--robot interaction requires robots whose actions are legible, allowing humans to interpret, predict, and feel safe around them. This study investigates the legibility of humanoid robot arm movements in a pointing task, aiming to understand how humans predict robot intentions from truncated movements and bodily cues. We designed an experiment using the NICO humanoid robot, where participants observed its arm movements towards targets on a touchscreen. Robot cues varied across conditions: gaze, pointing, and pointing with congruent or incongruent gaze.
Arm trajectories were stopped at 60\% or 80\% of their full length, and participants predicted the final target. 
We tested the multimodal superiority and ocular primacy hypotheses, both of which were supported by the experiment. 

\end{abstract}
\section{Introduction}
Smooth human--robot interaction (HRI) scenarios must involve robots that support human ability to interpret, predict, and feel safe around robotic actions \cite{Sciutti17}. Hence, the design of robotic motion must extend beyond efficiency in reaching a goal. One of the features critical for effective, safe, and explainable collaboration is the legibility of the robot trajectory, characterised by its distinctiveness that helps the observer disambiguate the robot intent \cite{Dragan13}. The legible movement supports transparency, which, together with human awareness, i.e., the robot ability to read and interpret human behavior, serves as a key pillar towards efficient HRI based on mutual understanding \cite{Mazzola25}.

To study human perception during HRI, it is essential to design robotic behavior to be repeatable and controllable \cite{Antonj23}. Thus, we designed an HRI experiment, using the humanoid robot NICO \cite{Kerzel17roman}, investigating the participants' ability to predict the intentions of robotic arm movement (reaching a point on a touchscreen) before completion. We propose a method to generate precise and controllable robotic arm trajectories \cite{Lucny25}. Combined with the pose of the robot head, these two modalities serve as information sources 
that human participants exploit in their inference task.
Based on this, we tested two hypotheses: 
\begin{itemize}
\item \textbf {Multimodal Superiority Hypothesis H\textsubscript{1}:} The accuracy of the target localisation significantly exceeds unimodal conditions when participants observed coherent 
gaze-pointing signals (GP), compared to gaze-only (G) and pointing-only (P) conditions.
\item \textbf {Oculomotor Primacy Hypothesis H\textsubscript{2}:} The reaction times (RTs) for gaze-only trials (RT\textsubscript{G}) are significantly shorter than for multimodal (RT\textsubscript{GP}) or pointing-only trials (RT\textsubscript{P}). 
\end{itemize}

\section{Related Work}
The ability of robots to communicate intent effectively is crucial for seamless HRI. Pointing gestures and gaze direction are fundamental deictic cues that humans instinctively use and interpret. Studies have shown that robot gaze can significantly direct human attention \cite{Admoni2017}, improve task efficiency and team fluency, and improve perceived social presence and engagement with the robot \cite{Kompatsiari2021}. Similarly, robotic pointing gestures are critical for disambiguating targets in a shared workspace, although their effectiveness can depend on factors such as embodiment and the clarity of the gesture itself \cite{Urakami2023} . 
Our work builds upon this foundation by examining the combined and individual effects of these cues specifically in the context of predicting intent from truncated arm movements.

The integration of multiple cues, such as simultaneous pointing and gazing, often leads to more robust and rapid understanding of a robot referent. Nonverbal cues, such as gaze and gesture, have been shown to enhance robot persuasiveness and clarity \cite{Chidambaram2012}. Research has also explored how humans resolve discrepancies when these cues are in conflict. For example, gaze has been shown to ``repair'' ambiguous pointing gestures, suggesting a potential hierarchy or differential weighting in how humans process these signals \cite{Mutlu2009}. More recent work investigates the nuances of these interactions, such as the timing and kinematics of combined gaze-gesture cues and their impact on perceived intentionality and trust. The current study contributes by systematically varying the congruency of movement and gaze and the completeness of the trajectories to understand how these factors modulate the interpretation of combined cues.

While much research has focused on completed actions, the interpretation of incomplete actions, where observers must extrapolate the robot goal, is less understood, particularly with multimodal cues. Anticipatory human responses to robot motions have been studied, showing that humans can predict robot intentions early in an action sequence \cite{Moon2014}. 
Our research extends this by quantifying the prediction accuracy for different percentages of trajectory completion (60\% vs. 80\%) across different combinations of gaze and pointing in order to delineate the thresholds at which robot intent becomes sufficiently legible.

\section{Materials and Methods}

\subsection {Environment}
The experiment was carried out using the NICO humanoid robot \cite{Kerzel17roman}
having 22 DoF in the head, shoulders, elbows, wrists, and fingers. 
The integrated hardware included two eye cameras, facial LED arrays for mouth and eyebrow expressions, and a built-in speaker.
The key interactive element was an LCD monitor with a capacitive touchscreen embedded in the table, serving as the robot target space. 
The experimental scene was further monitored by two external USB cameras that provided front and side views of the participant and the interaction. 
The robot behavior and experimental flow were managed by an integrated system of software agents communicating via a blackboard architecture, 
allowing modular control over various functions. 
Robot arm trajectories, consisting of 50 steps with 7 DoF, were precalculated using our novel methodology based on gradient descent and forward kinematics \cite{Lucny25} to ensure precise, linear, and repeatable movements from a consistent starting pose to selected targets on the touchscreen.
The source code for the experimental setup is publicly available.\footnote{\url{https://github.com/andylucny/nico2/tree/main/experiment}}

\begin{figure}[t!]
    \centering
    \includegraphics[width=1.0\textwidth]{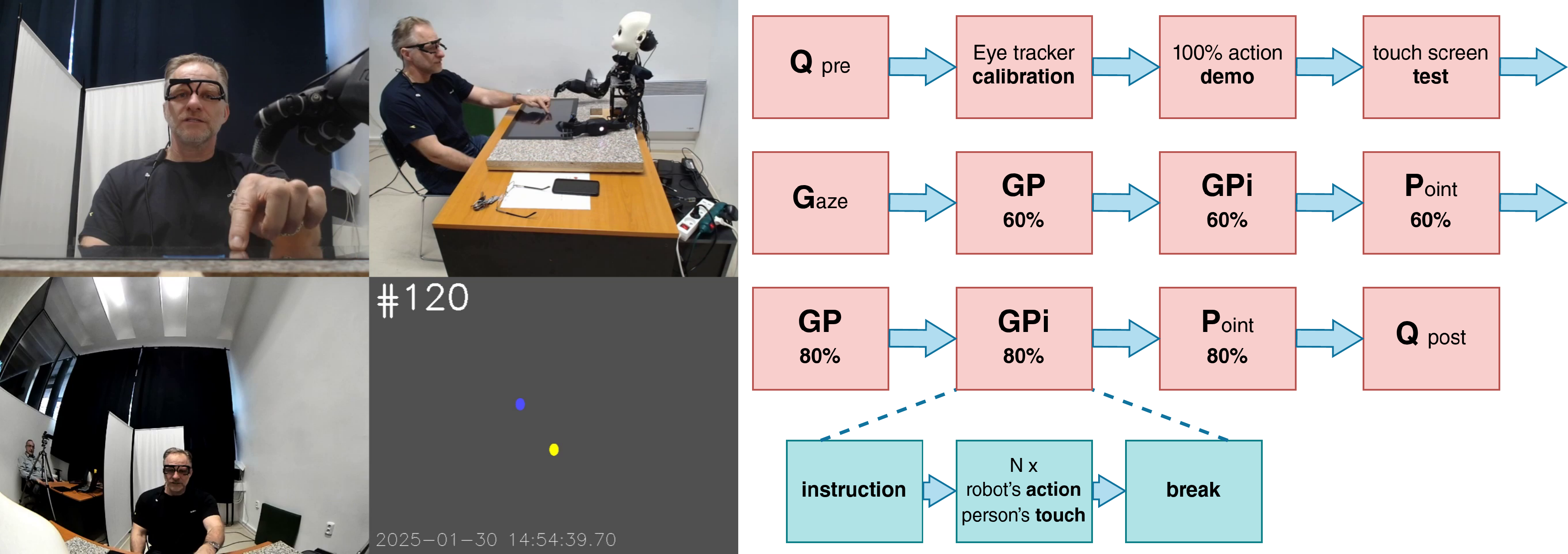}
    \caption{Left: Experimental setup as displayed in the software GUI - from left: camera 1 (top left), camera 2 (top right), the robot camera (bottom left), control screen with the blue dot representing the robot intention and the yellow dot representing the participant's prediction (bottom right). Right: general schema of the experimental procedure reflecting experimental batch~1.
    }
    \label{fig:schema}
\end{figure}

\subsection{Experimental Procedure}

The experiment started with pre-experiment questionnaires. In the main task, participants (students of various master's programs) predicted the target of NICO actions based on incomplete movements and social cues (gaze). NICO right arm trajectories were presented in two lengths: a shorter segment (60\%) and a longer segment (80\%) of the total distance to the target from the beginning.
Before the task began, the robot provided an initial verbal explanation of the task, along with a demonstration of a complete action that involved gaze and arm movement.

Four main experimental conditions were employed: a gaze-only (G) condition where the robot head was oriented towards a target; a pointing-only (P) condition where the robot executed arm movements; a pointing with the congruent gaze (GP) condition combining both cues towards the same target; and a pointing with the incongruent gaze (GPi) condition where the robot pointed at one target while gazing at a slightly shifted spot. 
Each participant experienced seven experimental sections: Since arm movements could provide rich information for the predictions, the order of conditions was set to gradually increase the informativeness of robot motor behavior (i.e. absence of arm movement, 60\% trajectory, 80\% trajectory). Moreover, the order of conditions was counterbalanced across two batches, when the information about the arm movement was equivalent:
\vspace{-0.5em}
\begin{itemize}
    \item {\bf Batch 1}: G, P60, GP60, GPi60, P80, GP80, GPi80;
    \item {\bf Batch 2}: G, GP60, GPi60, P60, GP80, GPi80, P80.
\end{itemize}
Each section was stated with an instruction from the robot, followed by five blocks consisting of seven randomized trials targeting one of the seven predefined points on the touchscreen (Fig.~\ref{fig:schema} left). During each trial, NICO performed its designated action (gazing, pointing, or both, see Fig.~\ref{fig:schema} right). An auditory beep then signaled that the participant would touch the screen at their predicted target location within two seconds. See the video at {\footnotesize\url{https://youtu.be/nqT8pidhHtc}}.


\subsection {Participants}
A total of 28 participants (11 male, 17 female) aged 18 to 35 years were recruited for the main study, 
to minimize potential variability in motor and sensory functioning that could confound the results. 
To maintain experimental control over linguistic variables,
all procedures were conducted exclusively in the Slovak language for native speakers.
Before the main data collection, a pilot study was conducted involving seven participants to validate the experimental procedure and its integrity. 

\subsection {Data Analysis}

\begin{figure}[t!]
    \centering
    \includegraphics[width=1.05\textwidth]{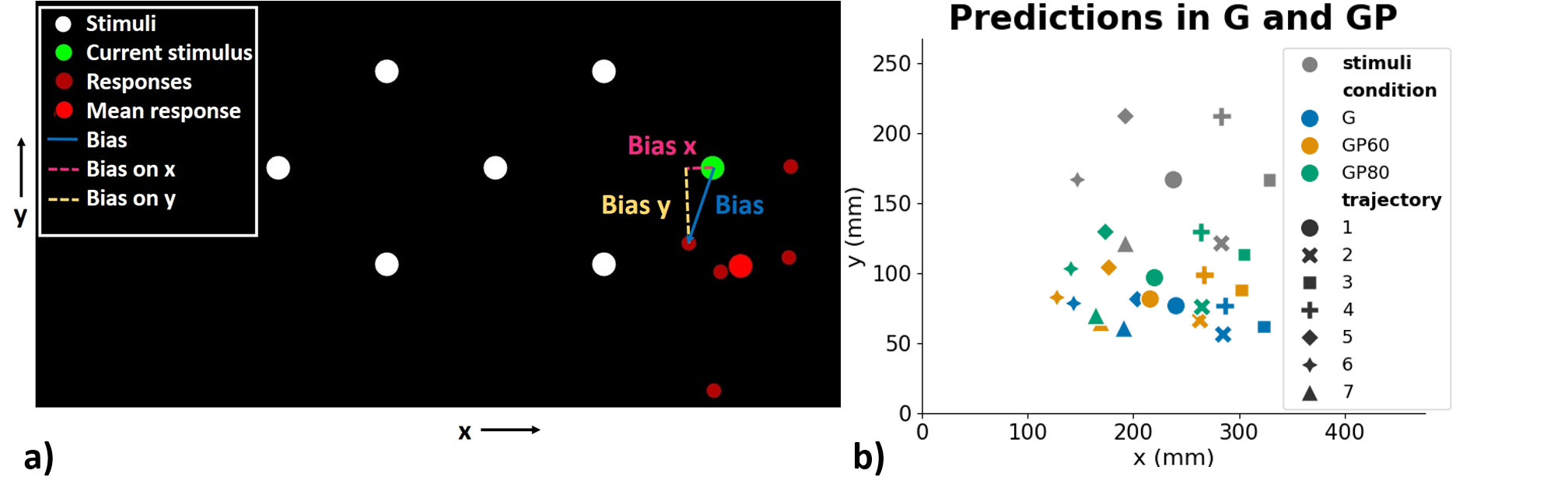}
    \vspace*{-10mm}
    \caption{a) Spatial layout of the seven target points (not shown to participants). The actual target is shown in green, participant responses in red, and remaining non-target options in white. b) Predictions in G and GP conditions. For each stimulus, mean predictions were computed for G, GP60 and GP80. The average distance between responses and stimuli identifies the bias.}
    \label{fig:targets}
\end{figure}

To evaluate the legibility of each experimental condition and to investigate the degree to which participants integrated gaze and pointing cues in the multimodal condition, we measured prediction accuracy through bias, calculated as the distance between the participant's touch point and the actual robot target. 
We decomposed bias into lateral components ($bias_{\rm x}$) and longitudinal ($bias_{\rm y}$), allowing us to quantify the magnitude and direction of the prediction error. 
From there, we derived the total bias for each target and condition as the mean Euclidean distance between the participants' responses and the corresponding target positions as
\begin{equation}
\label{eq:mean_bias}
bias_{\text{tot}} =\frac{1}{N}{\sum_{i=1}^{N} \sqrt{ \left( x_{\text{resp},i} - x_{\text{target},i} \right)^2 + \left( y_{\text{resp},i} - y_{\text{target},i} \right)^2}}
\end{equation}
where \( x_{\text{resp},i} \) and \( y_{\text{resp},i} \) are the coordinates of the participant's response in $i$-th trial, \( x_{\text{target},i} \) and \( y_{\text{target},i} \) are the corresponding coordinates of the actual target and $N$ is the number of times each target was presented.
Fig.~\ref{fig:targets} illustrates these measures in the spatial layout of the task on the touchscreen. It also displays the seven possible target locations.
The true target is highlighted in green, the responses of the participants in red (the larger is the mean), and the remaining non-target locations in white.


We measured participants' reaction time, defined as the time interval between the auditory cue indicating their response turn and the moment they touched the screen to provide their answer. Reaction times were collected for two key reasons. First, we instructed participants to respond as quickly as possible, promoting responses based on the ongoing motion trajectory rather than the final robot pose. Second, by comparing RTs across experimental conditions, our aim was to investigate how different communicative cues (gaze, pointing, and their combination) affected processing speed. 
Statistical analyses of bias and RT to test H\textsubscript{1}) and H\textsubscript{2}) were performed using linear mixed models \cite{gallucci2019} implemented in the Jamovi software\footnote{\url{https://www.jamovi.org/}}. 
The models included \texttt{condition} as a fixed effect, and both \texttt{participant\_ID} and \texttt{trajectory} as random intercepts to account for individual variability and target-specific effects (see eq.~\ref{eq:linearMixedModel}).
\begin{equation}
\label{eq:linearMixedModel}
\texttt{dep\_var} \sim 1 + \texttt{condition} + (1 | \texttt{participant\_ID}) + (1 | \texttt{trajectory})
\end{equation}

\section{Results}

\subsection*{Manipulation Check: Trajectory Legibility Improvement}

To verify the effectiveness of experimental manipulation of trajectory legibility, we performed a manipulation check to assess whether the trajectories designed to convey the robot intention with 80\% segment resulted in significantly lower prediction bias compared to those designed with 60\% segment. For this analysis, we used a paired-samples t-test, comparing the mean $bias_{\rm tot}$ for each participant. Before the analysis, we used the Shapiro-Wilk test, which did not indicate any significant deviation from normality ($W$ = 0.980, $p = .856$), supporting the suitability of the parametric test.
The paired-samples t-test revealed a significant reduction in total bias in the 80\% legibility condition ($M$=84.8mm, $SD$=21.9mm) compared to the 60\% condition ($M$=121mm, $SD$=29.2mm), $t$(26) = 8.78, $p < .001$. The mean difference was $M_{\text{diff}}$ = 36.2 mm, with a standard error of $SE$ = 4.12 mm. The effect size, measured by Cohen’s $d$, was large ($d = 1.69$), indicating a substantial improvement in the legibility of the trajectory.

\subsection*{Multimodal Superiority Hypothesis (H\textsubscript{1})}

\begin{figure}[ht]
  \centering
  \includegraphics[width=1.0\textwidth]{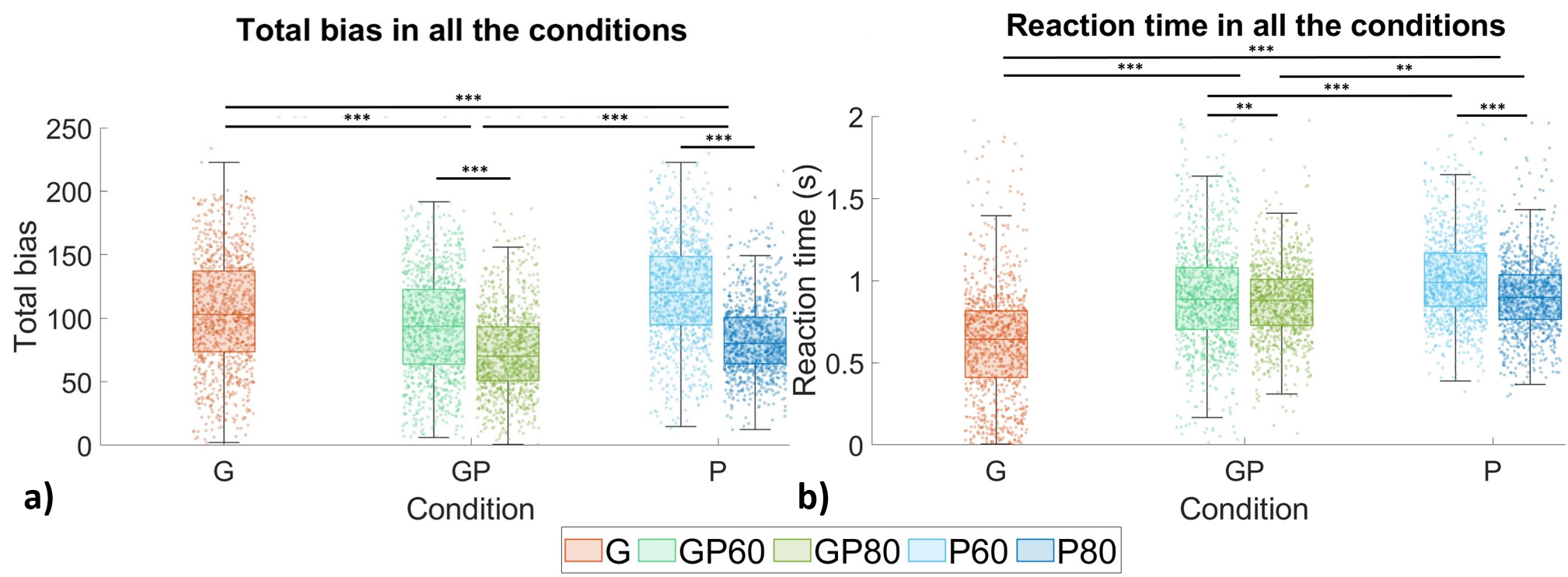}
  \vspace*{-6mm}
  \caption{a) Comparison of (a) total bias and (b) reaction times in five different conditions. Dots identify subjects' total bias (a) and reaction times (b) for each stimulus.  Horizontal bars above the graphs show significant differences, found through post hoc pairwise comparisons with Bonferroni correction.} 
  \label{fig:Bias and Reaction Time}
\end{figure}

To test Multimodal Superiority Hypothesis, we employed a Linear Mixed Model to assess differences in total bias across all experimental conditions (eq.~\ref{eq:linearMixedModel}). 
The model showed a significant effect of the condition on total bias (see 
Fig.~\ref{fig:targets}a to understand an example of one representative participants' error for one stimulus). 
The omnibus F-test for the fixed effect of condition was significant, $F(4, 5268) = 479$, $p < .001$, confirming that total bias significantly varied across conditions.
Post hoc pairwise comparisons with Bonferroni correction revealed that the legibility of gaze-only (G) trials was significantly higher (lower total bias) than in pointing-only trials with 60\% segment (P60; $t = -13.62$, $p < .001$), but lower than in pointing-only trials with 80\% segment (P80; $t$ = 17.21, $p < .001$). Importantly, when gaze and pointing cues were combined, participants’ performance improved significantly compared to both unimodal conditions (Fig.~\ref{fig:Bias and Reaction Time}a). This advantage was evident at both 60\% segment with a significant reduction of $bias_{\rm tot}$ (GP60-P60: $t = -23.17$, $p < .001$; GP60-G: $t = -9.47$, $p < .001$) and 80\% (GP80-P80: $t = -9.26$, $p < .001$; GP80-G: $t = -26.47$, $p < .001$), supporting the multimodal superiority hypothesis. As shown in Fig.~\ref{fig:targets}b, in gaze and pointing trials participants’ responses showed a lateral bias, in the opposite direction of the robot arm on the x-axis.  On the contrary, in gaze-only condition participants seemed to anchor their attention in the horizontal direction. Since on the y-axis participants' responses were characterized by a bias toward the robot body, the gaze seemed to have a role in mitigating prediction error, in particular on x-axis. 

\subsection*{Oculomotor Primacy Hypothesis (H\textsubscript{2})} 
The oculomotor primacy hypothesis was tested through a Linear Mixed Model to identify any differences in RT between all conditions (eq.~\ref{eq:linearMixedModel}). The model considered \texttt{condition} as a fixed effect, and both \texttt{participant\_ID} and \texttt{trajectory} as random intercepts to prevent individual variability and target-specific effects (adapted from Eq.~\ref{eq:linearMixedModel}).
According to the omnibus F-test for the fixed effect of condition, RTs significantly varied across conditions ($F(4, 5268) = 340$, $p < .001$).
Post hoc pairwise comparisons with Bonferroni correction showed that participants anticipated their response in G rather than P trials (G-P60: $t = -35.27$, $p < .001$; G-P80: $t = -25.32$, $p < .001$). Even when participants could combine G-P information, they were faster in G trials (G-GP60: $t = -25.38$, $p < .001$; G-GP80: $t = -21.62$, $p < .001$). Surprisingly, the combination of gaze and pointing allowed participants to have faster RTs, compared to P-only condition (P60-GP60: $t = 9.90$, $p < .001$; P80-GP80: $t = 3.71$, $p = .002$). The results shown in Fig.~\ref{fig:Bias and Reaction Time}b supported the oculomotor primacy hypothesis. 

\section{Discussion and Future Work}

Our initial results offer valuable insights into how robotic motion design and multimodal cues influence robot legibility \cite{Dragan13}. Our preliminary analysis illuminates the complementary roles of gaze and pointing cues. The Multimodal Superiority Hypothesis (H\textsubscript{1}) is supported by the implication that humans naturally integrate social (gaze) and spatial (pointing) information streams to form richer mental models of robot behavior \cite{Mazzola2023}. This integration facilitates a more robust understanding of the robot target, beyond what unimodal signals can provide.
In parallel, the support of the Oculomotor Primacy Hypothesis (H\textsubscript{2}) highlights gaze as an especially rapid and salient cue. The fact that gaze-only conditions yield the fastest responses suggests that gaze acts as an early attentional beacon, enabling swift orientation towards the robot goal. The finding that multimodal conditions also benefit from processing advantage of gaze further supports the idea that gaze effectively primes the observer’s perceptual system, streamlining the integration of additional cues.

The directional analysis of bias along the spatial axes provides additional insight. Participants’ responses consistently showed a lateral bias opposite to the arm direction along the x-axis, indicating that the participants tended to predict the motion endpoint to extend beyond the actual target. Interestingly, no such lateral bias was observed in gaze-only conditions, suggesting that gaze cues helped anchor attention precisely along the horizontal plane. Along the longitudinal, y-axis, a bias toward the robot body was evident, indicating a tendency to underestimate the forward reach distance. This pattern suggests that gaze cues help mitigate the prediction bias, in particular along the x-axis.

One of the future prospects of this work lies in developing a Bayesian model that quantifies how participants integrate gaze and pointing signals. 
Such a model could clarify the relative contribution of each cue to perception and decision-making, 
highlighting important implications for designing robotic systems that can communicate their intentions effectively and foster shared understanding in mixed human--robot environments.

\begin{credits}
\subsubsection{\ackname} This research has received funding from the Horizon Europe project TERAIS, the Grant agreement no.~101079338.
\end{credits}
%

\end{document}